\documentclass{article}

\usepackage{PRIMEarxiv}
\usepackage{natbib}
\usepackage{CJKutf8}
\usepackage{CJK}
\usepackage[utf8]{inputenc} 
\usepackage[T1]{fontenc}    
\usepackage{hyperref}       
\usepackage{url}            
\usepackage{booktabs}       
\usepackage{amsfonts}       
\usepackage{nicefrac}       
\usepackage{microtype}      
\usepackage{lipsum}
\usepackage{graphicx}
\usepackage{amsmath}
\usepackage{pifont}
\usepackage[inline]{enumitem}
\usepackage{enumitem}
\usepackage{boldline}
\usepackage{hhline}
\usepackage{multirow}
\usepackage{xcolor}
\usepackage{fontawesome5}
\usepackage{tablefootnote} 
\usepackage{listings}
\usepackage{booktabs,threeparttable,siunitx,xcolor,pgfplots}
\usepackage{graphicx}
\usepackage{xspace} 

\usepackage[most]{tcolorbox}
\usepackage{enumitem}
\usepackage{xspace}

\newtcolorbox{sysbox}[1][]{
  colback=gray!6,
  colframe=black!20,
  boxrule=0.5pt,
  arc=1.5mm,
  left=1mm,right=1mm,top=1mm,bottom=1mm,
  #1
}

\newtcolorbox{promptbox}[2][]{%
  enhanced,
  breakable,
  colback=black!2,
  colframe=black!70,
  boxrule=0.7pt,
  arc=2pt,
  left=6pt,right=6pt,top=6pt,bottom=6pt,
  title={#2},
  fonttitle=\bfseries,
  #1
}

\newtcolorbox{overviewbox}{
  colback=gray!6,
  colframe=black!20,
  boxrule=0.5pt,
  arc=1.2mm,
  left=1.2mm,right=1.2mm,top=1mm,bottom=1mm
}

\newtcolorbox{layoutbox}[1]{
  colback=black!2,
  colframe=black!18,
  boxrule=0.45pt,
  arc=1mm,
  left=1.5mm,right=1.5mm,top=1mm,bottom=1mm,
  title=\textbf{#1},
  fonttitle=\small,
  coltitle=black
}

\definecolor{skillblue}{HTML}{2F5D8A}
\definecolor{skilllight}{HTML}{F4F8FC}
\definecolor{skillborder}{HTML}{A9C3DE}
\definecolor{tagbg}{HTML}{EAF2FB}

\newtcbox{\skilltag}{
  on line,
  boxrule=0pt,
  arc=2mm,
  colback=tagbg,
  colframe=tagbg,
  left=2mm,right=2mm,top=0.6mm,bottom=0.6mm
}

\setlist[itemize]{leftmargin=1.5em, topsep=2pt, itemsep=2pt, parsep=0pt}
\setlist[enumerate]{leftmargin=1.8em, topsep=2pt, itemsep=2pt, parsep=0pt}
\setlist[description]{leftmargin=1.6em, style=nextline, topsep=2pt, itemsep=3pt}

\usepackage[utf8]{inputenc}
\graphicspath{{images/}}     
\usepackage{amsthm}
\theoremstyle{definition}

\usepackage[most]{tcolorbox}

\usepackage[most]{tcolorbox}
\usepackage{verbatim} 

\newtcolorbox{findingbox}{
  colback=black!5!white,      
  colframe=black!75!black,    
  arc=3mm,                    
  boxrule=0.5pt,              
  left=4mm, right=4mm, top=3mm, bottom=3mm 
}
\usepackage[most]{tcolorbox}

\usepackage[most]{tcolorbox}

\newtcolorbox{caselogbox}[1]{
  prompt_box1_style, 
  title=#1          
}

\newtcolorbox{promptlogbox}[1]{
  prompt_box2_style, 
  title=#1          
}

\usepackage{tcolorbox}
\usepackage{listings} 
\tcbuselibrary{breakable, listings} 

\tcbuselibrary{listings} 
\usepackage[most]{tcolorbox}

\newtcolorbox{textBox}[1][]{
  colback=gray!5,
  colframe=gray!60!black,
  fonttitle=\bfseries,
  colbacktitle=gray!85!black,
  title=#1,
  breakable
}

\title{AutoSkill: Experience-Driven Lifelong Learning via Skill Self-Evolution}

\author{
  \textbf{Yutao Yang}$^{1}$\footnotemark[1], \textbf{Junsong Li}$^1$\footnotemark[1], Qianjun Pan$^{1}$\footnotemark[1], \textbf{Bihao Zhan}$^{1}$, \textbf{Yuxuan Cai}$^{1}$, \textbf{Lin Du}$^{1}$, \textbf{Jie Zhou}$^{1,2}$\footnotemark[2], \textbf{Kai Chen}$^2$\footnotemark[2], \\ \textbf{Qin Chen}$^1$, \textbf{Xin Li}$^2$, \textbf{Bo Zhang}$^2$, \textbf{Liang He}$^1$ \\
  $^1$ School of Computer Science and Technology, East China Normal University, $^2$ Shanghai AI Laboratory\\ 
  \texttt{\{jzhou, qchen, lhe\}@cs.ecnu.edu.cn}, \\
  \textcolor{red}{\url{https://github.com/ECNU-ICALK/AutoSkill}} 
}

\begin{document}

\renewcommand{\thefootnote}{\fnsymbol{footnote}}

\maketitle

\footnotetext[1]{These authors contributed equally to this work.}
\footnotetext[2]{Corresponding Authors}

\begin{abstract}
In practical LLM applications, users repeatedly express stable preferences and requirements—such as reducing hallucinations, following institutional writing conventions, or avoiding overly technical wording—yet such interaction experience is seldom consolidated into reusable knowledge. Consequently, LLM agents often fail to accumulate personalized capabilities across sessions. We present AutoSkill, an experience-driven lifelong learning framework that enables LLM agents to automatically derive, maintain, and reuse skills from dialogue and interaction traces.

AutoSkill abstracts skills from user experience, supports their continual self-evolution, and dynamically injects relevant skills into future requests without retraining the underlying model. Designed as a model-agnostic plug-in layer, it is compatible with existing LLMs and introduces a standardized skill representation for sharing and transfer across agents, users, and tasks. In this way, AutoSkill turns ephemeral interaction experience into explicit, reusable, and composable capabilities.

This paper describes the motivation, architecture, skill lifecycle, and implementation of AutoSkill, and positions it with respect to prior work on memory, retrieval, personalization, and agentic systems. AutoSkill highlights a practical and scalable path toward lifelong personalized agents and personal digital surrogates.
\end{abstract}


\section{Introduction}

Large language models \cite{brown2020language,touvron2023llama,deepseekai2025deepseekr1incentivizingreasoningcapability} have enabled a new generation of interactive agents for writing assistance, planning, coding, decision support, and tool use~\cite{yao2022react,schick2023toolformer,patil2024gorilla,park2023generative}. As these systems move from controlled benchmarks to real world deployment, a recurring pattern becomes increasingly visible: users repeatedly restate stable preferences and operating requirements across sessions. For example, a user may consistently ask the agent to avoid hallucinations, follow an official writing style, reduce technical jargon, or adhere to a preferred workflow. Recent work on memory enhanced agents and long horizon conversational settings has highlighted the importance of preserving user specific information over time~\cite{zhong2024memorybank,packer2023memgpt,maharana2024evaluating,wu2024longmemeval,chhikara2025mem0}. However, in current practice, such repeated interaction experience is rarely transformed into reusable capability. As a result, user habits and task specific expectations often need to be reestablished from scratch in each new session.

This limitation reveals a broader challenge for personalized language agents. Existing approaches provide only partial solutions. Parameter updating and self evolution methods can improve model behavior through self reflection, feedback driven optimization, or self training~\cite{lu2023self,huang2023large,qu2024recursive,wang2025self}, but they are often costly or difficult to control in settings that require frequent and fine grained personalization. Memory based approaches preserve facts, preferences, or prior dialogue content through retrieval and long term storage~\cite{lewis2020retrieval,zhong2024memorybank,packer2023memgpt,chhikara2025mem0,xu2025mem,salama2025meminsight}, yet they usually treat past interaction as text to be retrieved rather than behavior to be operationalized. Agent frameworks and skill learning methods have demonstrated the value of reusable strategies for reasoning, tool use, and task execution~\cite{yao2022react,schick2023toolformer,wang2023voyager,shinn2023reflexion}, but in many cases those skills remain implicit in prompts, trajectories, or policies. What is still missing is a mechanism that can convert recurring interaction experience into explicit, reusable, and maintainable skills.

In this paper, we present \textsc{AutoSkill}, an experience driven lifelong learning framework for large language model agents. The central idea of AutoSkill is to treat repeated interaction experience not merely as memory, but as a source of skill formation. Instead of storing only dialogue snippets or preference records, AutoSkill abstracts reusable behaviors from user interactions and crystallizes them into explicit skill artifacts. These artifacts capture behavioral patterns such as stylistic constraints, response strategies, tool use procedures, and domain specific operating conventions. Because they are represented in a structured form, skills can be inspected, edited, merged, versioned, and reused across sessions.

AutoSkill supports a full skill lifecycle. It identifies candidate skills from dialogue and interaction events, summarizes them into standardized \texttt{SKILL.md} artifacts, updates them through iterative refinement, and injects relevant skills into future requests at inference time. This design enables continual capability accumulation without retraining the underlying model. It also provides a practical interface for human oversight, since developers and users can directly inspect and revise the resulting skills. In this way, AutoSkill bridges short term interaction experience and long term capability development, moving language agents closer to the goal of becoming personal digital surrogates that reflect stable user habits, preferences, and working styles.

Beyond practical personalization, AutoSkill contributes a distinct perspective on lifelong learning for language agents. It shifts the unit of accumulation from raw memory records to explicit behavioral knowledge, and it frames agent improvement as a process of skill extraction, maintenance, and reuse. This perspective is important for both research and deployment. From a research standpoint, it offers a concrete representation for studying how interaction experience can become reusable capability. From a system standpoint, it provides a plug in layer that can work with existing language models and agent frameworks, while supporting skill sharing and transfer across tasks and users.

The main contributions of this paper are as follows:
\begin{itemize}[leftmargin=*, align=left] 
    \item We formulate the problem of transforming interaction experience into explicit reusable skills for personalized large language model agents, and we introduce AutoSkill as a framework for this setting.
    \item We propose a skill lifecycle that covers skill extraction, structured representation, iterative refinement, retrieval, and reuse, enabling continual skill evolution without modifying base model parameters.
    \item We design skills as editable and versioned artifacts, which improves transparency, controllability, and long term maintainability compared with implicit memory or policy based approaches.
    \item We implement AutoSkill as an open source and deployable system that supports integration with existing large language models and agent pipelines, providing a practical path toward lifelong personalized agents.
\end{itemize}

\section{Related Work}

We organize related work into four research threads: lifelong learning from experience, self evolution for large language models, long term memory for language agents, and skill learning for reasoning and acting. AutoSkill is related to all four directions, but it is distinguished by its emphasis on explicit skill artifacts, human editable representations, and lifecycle management for skill extraction, revision, retrieval, and reuse.

\subsection{Experience Driven Lifelong Learning}

Experience driven lifelong learning studies how agents accumulate reusable knowledge, strategies, or policies from ongoing interactions, so that experience obtained in one setting can support performance in future tasks. Core questions in this line of work include when knowledge should be extracted from interaction history, what should be retained as reusable capability, and how noise accumulation and forgetting can be controlled over long horizons. Experience driven Lifelong Learning (ELL) formalizes these goals and introduces benchmarks such as StuLife for evaluating self evolving agents in long horizon environments~\cite{cai2025building}. Related surveys further organize the research landscape around the perception, memory, and action pipeline of lifelong LLM agents~\cite{zheng2026lifelong}. AutoSkill shares the goal of continual capability accumulation from real interactions, but differs in how experience is represented and maintained. Instead of keeping knowledge as latent memory or implicit policy adaptation, AutoSkill crystallizes reusable capabilities into explicit \texttt{SKILL.md} artifacts with versioned evolution. This design improves interpretability, supports manual inspection and revision, and makes sustained alignment with user preferences easier to achieve.

\subsection{Self Evolution for Large Language Models}

Self evolution methods aim to improve model behavior through self reflection, iterative rewriting, feedback driven refinement, or autonomous data construction. Representative work includes SELF, which introduces self evolution with language feedback~\cite{lu2023self}; \emph{Large Language Models Can Self Improve}, which studies self training with unlabeled data~\cite{huang2023large}; and Recursive Introspection (RISE), which enables models to revise prior attempts through repeated introspection~\cite{qu2024recursive}. Self Evolving Curriculum (SEC) further explores automated curriculum construction for reasoning tasks~\cite{chen2025self}, while recent surveys summarize the broader landscape of self evolving LLM systems~\cite{tao2024survey}. Uncertainty enhanced preference optimization (UPO) is another representative approach, where model policies are improved through reliable feedback sampling~\cite{wang2025self}. AutoSkill is complementary to this line of research. It does not update model parameters or rely on implicit policy drift. Instead, it externalizes reusable behaviors into structured skill artifacts and supports their evolution through explicit revision, merging, and version control. This makes the improvement process more transparent and controllable, especially in scenarios where user preferences and working styles must remain stable across sessions.

\subsection{Long Term Memory for Language Agents}

Retrieval augmented generation (RAG) improves factuality and traceability by injecting retrieved external knowledge into the generation process~\cite{lewis2020retrieval}. Retrieval augmented pretraining and retrieval enhanced language models, including REALM~\cite{guu2020retrieval} and RETRO~\cite{borgeaud2022improving}, extend this idea by coupling parametric models with large non parametric memory. Dense and unsupervised retrieval methods such as DPR~\cite{karpukhin2020dense} and Contriever~\cite{izacard2021unsupervised}, as well as late interaction retrievers such as ColBERT~\cite{khattab2020colbert}, improve retrieval quality and efficiency. Fusion based methods including FiD and Atlas further demonstrate the effectiveness of retrieval augmented reasoning for question answering~\cite{izacard2021leveraging,izacard2023atlas}. Other approaches, such as kNN LM, extend model recall through nearest neighbor retrieval in representation space~\cite{khandelwal2019generalization}. Beyond factual retrieval, memory oriented systems introduce mechanisms for long term storage and management across sessions, including MemoryBank~\cite{zhong2024memorybank}, MemGPT~\cite{packer2023memgpt}, and generative agent architectures that organize episodic memories and reflections for planning~\cite{park2023generative}. Benchmarks such as LoCoMo~\cite{maharana2024evaluating}, LongMemEval~\cite{wu2024longmemeval}, and incremental multi turn memory evaluation~\cite{hu2025evaluating} assess long horizon memory in conversational agents. More recent frameworks, including Mem0~\cite{chhikara2025mem0}, A MEM~\cite{xu2025mem}, and MemInsight~\cite{salama2025meminsight}, together with corresponding surveys~\cite{zhang2025survey}, further systematize memory mechanisms for LLM based agents. AutoSkill builds on the insight that retrieval can reactivate useful past experience, but it moves beyond conventional memory systems by lifting memory from text records to behavior units. Through explicit skill abstraction, retrieval, and maintenance, AutoSkill is better suited for preserving stable preferences, stylistic constraints, and recurring workflows that are difficult to represent as raw text snippets alone.

\subsection{Skill Learning for Reasoning and Acting Agents}

Skill learning for LLM agents concerns the acquisition of reusable reasoning patterns, tool use procedures, and action strategies. Methods for agentic reasoning and decision making, such as ReAct~\cite{yao2022react}, show that interleaving reasoning and acting can improve tool interactive problem solving. Tool use oriented approaches, including Toolformer~\cite{schick2023toolformer}, ART~\cite{paranjape2023art}, ToolAlpaca~\cite{tang2023toolalpaca}, and Gorilla~\cite{patil2024gorilla}, further show that language models can learn to invoke external tools and APIs in increasingly general settings. Related benchmarks and datasets, such as API Bank~\cite{li2023api} and ToolBench or ToolLLM~\cite{qintoolllm}, provide evaluation settings for tool use competence. In embodied and open ended environments, systems such as Voyager highlight the value of compositional skill libraries for continual exploration and reuse~\cite{wang2023voyager}. Reflexion also uses verbal feedback and memory updates to improve future decisions~\cite{shinn2023reflexion}. A range of agent benchmarks and environments, including WebShop~\cite{yao2022webshop}, ALFWorld~\cite{shridharalfworld}, WebArena~\cite{zhou2023webarena}, and AgentBench~\cite{liu2023agentbench}, further stress long horizon execution, planning, and skill generalization. However, in most existing approaches, skills remain implicit in prompts, trajectories, or latent policies, and therefore lack a unified mechanism for inspection, editing, transfer, and long term maintenance. AutoSkill addresses this limitation by treating skills as first class artifacts that can be extracted from interaction experience, edited by users or developers, merged across iterations, versioned over time, and dynamically injected into future tasks. This explicit extraction and maintenance loop is central to AutoSkill and enables sustained skill evolution in a controllable manner.

\begin{figure}[t!]
    \centering
    \includegraphics[width=1.0\textwidth]{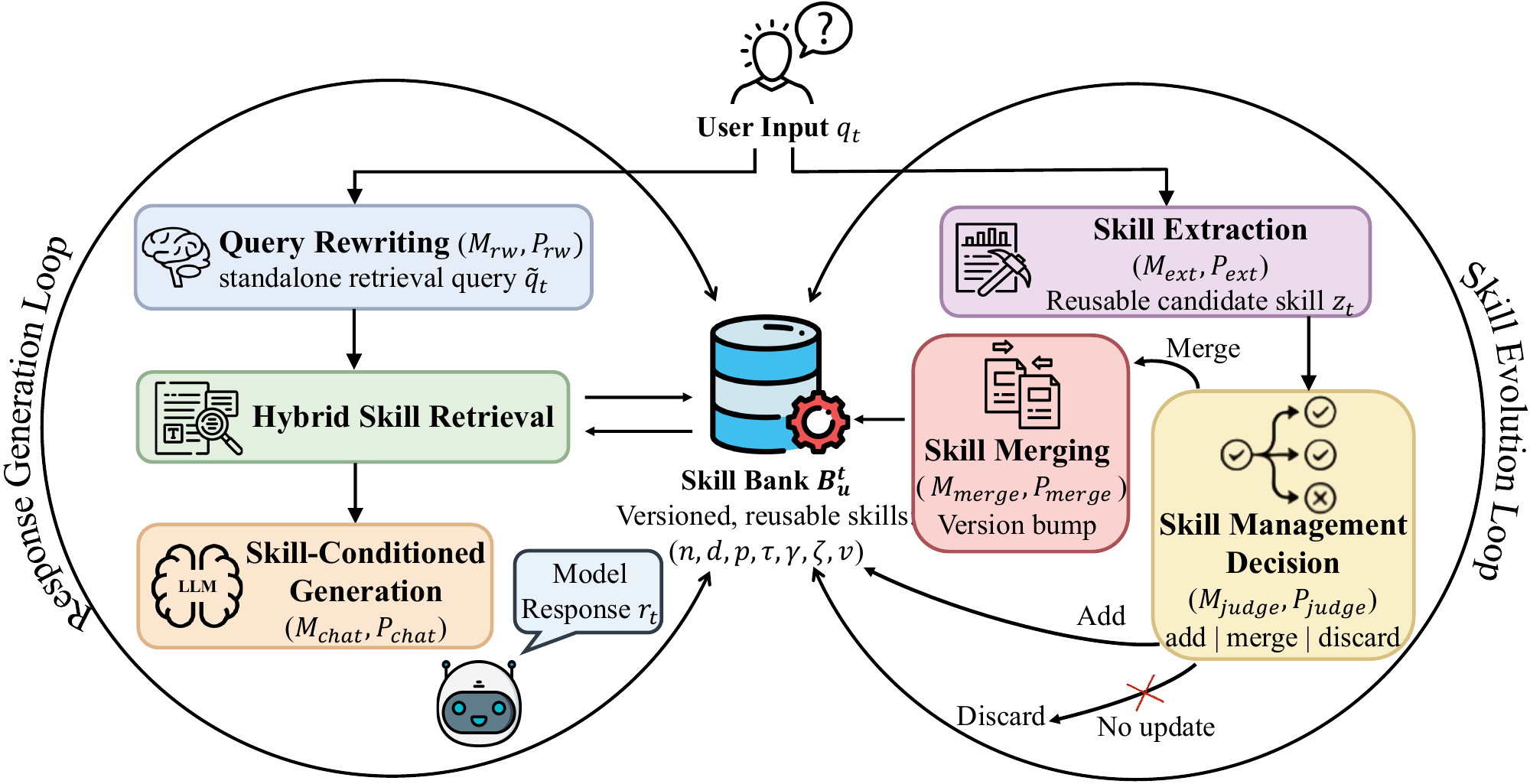}
    \caption{The Framework for our AutoSkill, which is composed of two tightly coupled processes. The right loop, \emph{skill evolution}, transforms interaction experience into explicit skills through extraction and maintenance. The left loop, \emph{skill-enhanced response generation}, uses the current skill bank to support response generation via query rewriting, skill retrieval, and context injection. In this way, the system continually improves through explicit memory growth rather than through model fine-tuning. }
    \label{fig:framework}
\end{figure}

\section{Method}

We propose a training-free lifelong learning framework that improves dialogue quality through explicit skill self-evolution rather than parameter updates. The core idea is to externalize reusable task-solving patterns as versioned skills, retrieve them for future responses, and continuously refine them with newly observed user interactions. As shown in Figure~\ref{fig:framework}, the framework consists of two coupled loops: \emph{skill-enhanced response generation} and \emph{skill evolution}. The former retrieves useful skills to assist the current response, while the latter updates the skill bank based on newly observed dialogue turns.

\subsection{Problem Definition}

For a user $u$, we denote the complete dialogue history as
\[
\mathcal{X}_u = \{x_1, x_2, \dots, x_T\}, \qquad x_t = (q_t, r_t),
\]
where $q_t$ is the user query at turn $t$ and $r_t$ is the model response. A skill bank $\mathcal{B}_u^t$ is maintained for user $u$ after turn $t$. Each skill is represented as
\[
s = (n, d, p, \tau, \gamma, \xi, v),
\]
where $n$ is the skill name, $d$ is the description, $p$ is the executable instruction prompt, $\tau$ is the trigger set, $\gamma$ is the tag set, $\xi$ is the example set, and $v$ is the version.

Our method is \textbf{training-free}: no model parameters are updated during deployment. Instead, the system is implemented with five prompt-driven modules: a query rewriting model, a dialogue response model, a skill extraction model, a skill management judge, a skill merge model, and an embedding model for skill vectorization. Given the current query $q_t$ and dialogue history, the system retrieves relevant skills from $\mathcal{B}_u^t$ to generate the response $r_t$; meanwhile, it uses the \emph{user-side} interaction signal to update the skill bank. Importantly, the skill extraction stage only uses user queries rather than model responses, i.e., it learns from $\{q_1,\dots,q_t\}$ but does not use $r_t$ as extraction evidence.

\subsection{Prompt-Driven Modular Architecture}

All functional modules in our framework are realized by task-specific prompts rather than specialized training. Let
\[
\mathcal{P} = \{P_{\mathrm{rw}}, P_{\mathrm{chat}}, P_{\mathrm{ext}}, P_{\mathrm{judge}}, P_{\mathrm{merge}}\}
\]
denote the prompt set for query rewriting, dialogue generation, skill extraction, skill management decision, and skill merging, respectively. Each module is instantiated by pairing a general-purpose LLM with its corresponding prompt. Therefore, our framework can be viewed as a modular inference-time composition:
\[
\mathcal{M} = \{M_{\mathrm{rw}}, M_{\mathrm{chat}}, M_{\mathrm{ext}}, M_{\mathrm{judge}}, M_{\mathrm{merge}}, M_{\mathrm{emb}}\},
\]
where $M_{\mathrm{emb}}$ is the embedding model used for dense vector retrieval.

This design has two advantages. First, different modules can share the same backbone LLM while serving different roles through different prompts. Second, the whole system remains highly flexible: replacing the response model, extraction model, or embedding model does not require retraining the framework itself.

\subsection{Skill-Enhanced Response Generation}

Given the current user query $q_t$ and recent dialogue history $h_t \subset \mathcal{X}_u$, the system first rewrites the query into a retrieval-oriented form by a dedicated LLM prompt:
\[
\tilde{q}_t = M_{\mathrm{rw}}(P_{\mathrm{rw}}, q_t, h_t).
\]
The purpose of query rewriting is to resolve context dependence, preserve the current task anchor, and expose retrieval-critical constraints such as format, style, structure, or domain requirements.

\begin{promptbox}{\boldmath$P_{\mathrm{rw}}$ : Query Rewriting}
\textbf{Role.} Retrieval query rewriter.

\textbf{Goal.} Rewrite the current user input into exactly one concise,
standalone search query for skill retrieval.

\textbf{Core rules.}
\begin{itemize}[leftmargin=1.2em, nosep]
  \item First judge whether the current turn is a continuation of the same task
  or a topic switch.
  \item For continuation, preserve the existing \emph{topic anchor} and append
  only the new constraints.
  \item For topic switch, replace the previous anchor with the new task/topic.
  \item If the current turn contains only style/format constraints, inherit the
  missing task anchor from recent history.
  \item Resolve references such as ``it'', ``this'', and ``the above''.
  \item Keep only retrieval-relevant constraints (e.g., format, audience,
  quality requirements, banned structures).
  \item Avoid generic process words without a concrete task/topic anchor.
\end{itemize}

\textbf{Output.} One-line rewritten query, in the same language as the user.
\end{promptbox}

\subsubsection{Hybrid Skill Retrieval}
For each skill $s \in \mathcal{B}_u^t$, we compute both a dense semantic relevance score and a lexical BM25 relevance score:
\[
d(q_t, s) = \mathrm{sim}\!\left(M_{\mathrm{emb}}(\tilde{q}_t), M_{\mathrm{emb}}(s)\right),
\]
\[
b(q_t, s) = \mathrm{BM25}(\tilde{q}_t, s).
\]
Since these two scores lie on different scales, we normalize them into $[0,1]$ and combine them by weighted summation:
\[
\mathrm{Rel}(q_t, s) = \lambda \,\hat{d}(q_t, s) + (1-\lambda)\,\hat{b}(q_t, s),
\]
where $\lambda \in [0,1]$ controls the trade-off between dense semantic matching and lexical exact matching.

We then rank all skills by $\mathrm{Rel}(q_t, s)$ and keep only the top-$K$ candidates whose score exceeds a predefined threshold $\eta$:
\[
\mathcal{H}_t = \left\{ s \in \mathrm{TopK}(\mathcal{B}_u^t) \mid \mathrm{Rel}(q_t, s) \ge \eta \right\}.
\]
Only skills in $\mathcal{H}_t$ are injected into the dialogue model. If no skill satisfies the threshold, the model responds without skill augmentation.

\subsubsection{Skill-conditioned Response Generation}
The selected skills are rendered as a compact external memory context
\[
C_t = \mathrm{Render}(\mathcal{H}_t),
\]
and appended to the response prompt. The final response is generated by
\[
r_t = M_{\mathrm{chat}}(P_{\mathrm{chat}}, q_t, h_t, C_t).
\]
This makes the response model adaptive to user-specific accumulated experience while keeping the model parameters unchanged.

\begin{promptbox}{\boldmath$P_{\mathrm{chat}}$ : Dialogue Generation}
\textbf{Role.} Response generator with retrieved skill context.

\textbf{System policy.}
\begin{itemize}[leftmargin=1.2em, nosep]
  \item Behave as a helpful assistant.
  \item Retrieved skills may be irrelevant.
  \item Use a skill only when it directly matches the user's current intent.
  \item Otherwise, ignore all retrieved skills and answer normally.
  \item Never explicitly mention that such skills were retrieved/injected.
\end{itemize}

\textbf{Injected context template.}
\begin{quote}\small
Retrieved skill list \\
Search query: $\langle q \rangle$ \\
For each skill: \{name, id, description, tags, triggers, prompt\}
\end{quote}

\textbf{User-side input template.}
\begin{quote}\small
Conversation history: $\langle h \rangle$ \\
Respond to the latest user message.
\end{quote}
\end{promptbox}
\subsection{Real-Time Skill Evolution}
\subsubsection{Skill Extraction from Interaction}

After turn $t$, the framework attempts to induce a reusable skill candidate from user-side interaction signals. Since the purpose of skill extraction is to capture stable user requirements rather than model-generated content, we only use user queries as extraction evidence. Let
\[
\mathcal{Q}_{u}^{\,t} = \{q_1, q_2, \dots, q_t\}
\]
denote the user-query sequence up to turn $t$. The extraction module operates on a recent window of user queries:
\[
z_t = M_{\mathrm{ext}}(P_{\mathrm{ext}}, \mathcal{Q}_{u}^{\,t}),
\]
where $z_t$ is a skill candidate of the form
\[
z_t = (n, d, p, \tau, \gamma, \xi, c),
\]
with $c$ being the confidence score.

\begin{promptbox}{\boldmath$P_{\mathrm{ext}}$ : Skill Extraction}
\textbf{Role.} Skill extractor that turns conversation/event traces into reusable skills.

\textbf{Extraction principles.}
\begin{itemize}[leftmargin=1.2em, nosep]
  \item Treat user turns as the primary evidence; assistant turns are only context.
  \item Focus on recent rounds and detect boundary turns to avoid mixing different tasks.
  \item Extract only when there are durable, reusable constraints, policies,
  workflows, or templates.
  \item Do \emph{not} extract one-shot requests, generic tasks, stale constraints,
  or assistant-invented details.
  \item Capture \emph{how to do} similar tasks, rather than this-instance facts.
  \item Remove case-specific entities and preserve only portable rules.
  \item Do not invent workflow steps unless the user explicitly specified them.
\end{itemize}

\textbf{Output schema.}
\begin{quote}\small
\{"skills": [skill$_1$, \ldots, skill$_k$]\}
\end{quote}

\textbf{Skill fields.}
\begin{itemize}[leftmargin=1.2em, nosep]
  \item \texttt{name}: concise, searchable, intent-explicit;
  \item \texttt{description}: what the skill does and when to use it;
  \item \texttt{prompt}: Markdown body with
  \texttt{\# Goal}, \texttt{\# Constraints \& Style}, and optional
  \texttt{\# Workflow};
  \item \texttt{triggers}, \texttt{tags}, \texttt{examples}, \texttt{confidence}.
\end{itemize}
\end{promptbox}

The extraction prompt is designed to identify \emph{reusable} and \emph{durable} knowledge, such as persistent preferences, reusable procedures, output constraints, task-specific policies, or recurring corrections. In contrast, one-off requests or transient content should not be extracted as skills. Therefore, extraction serves as a structured abstraction process from raw user queries to reusable capability units.

\subsubsection{Retrieval-Assisted Skill Management}

A newly extracted candidate $z_t$ is not directly written into the skill bank. Instead, the system first retrieves the most similar existing skills and uses them as local evidence for maintenance decisions. This avoids feeding the entire skill bank into the judge or merge module.

Specifically, the candidate $z_t$ is converted into a retrieval query based on its name, description, triggers, and instructions. Similar to online response retrieval, we compute a hybrid relevance score between $z_t$ and each existing skill $s \in \mathcal{B}_u^t$:
\[
\mathrm{Rel}_{\mathrm{m}}(z_t, s) = \alpha \,\hat{d}(z_t, s) + (1-\alpha)\,\hat{b}(z_t, s),
\]
where $\alpha$ is the management-time retrieval weight. We then retrieve a small neighbor set
\[
\mathcal{N}_t = \mathrm{TopM}\left(\mathcal{B}_u^t; \mathrm{Rel}_{\mathrm{m}}(z_t, s)\right),
\]
and select the most similar existing skill
\[
s_t^\ast = \arg\max_{s \in \mathcal{N}_t} \mathrm{Rel}_{\mathrm{m}}(z_t, s).
\]

The management decision is then made by a dedicated prompt-driven judge:
\[
a_t = M_{\mathrm{judge}}(P_{\mathrm{judge}}, z_t, s_t^\ast),
\qquad
a_t \in \{\texttt{add}, \texttt{merge}, \texttt{discard}\}.
\]
In other words, the judge only needs to compare the current candidate with its most relevant memory neighbor, rather than reasoning over the whole skill bank. This makes the decision process both more focused and more scalable.

\begin{promptbox}{\boldmath$P_{\mathrm{judge}}$ : Skill Management Decision}
\textbf{Role.} Skill set manager for deciding \texttt{add}, \texttt{merge}, or \texttt{discard}.

\textbf{Decision procedure.}
\begin{enumerate}[leftmargin=1.4em, nosep]
  \item Check topic continuity and capability family:
  whether the candidate continues the same ongoing work item.
  \item Apply a discard gate:
  reject generic, low-signal, non-portable, or library-covered candidates.
  \item Compare candidate vs.\ existing user skills on four axes:
  job-to-be-done, deliverable type, hard constraints/success criteria,
  and required tools/workflow.
  \item Choose \texttt{merge} only when they are the same capability after
  removing instance details.
  \item Choose \texttt{add} when the candidate remains a distinct durable capability.
\end{enumerate}

\textbf{Output schema.}
\begin{quote}\small
\{"action": add|merge|discard,\; "target\_skill\_id": id|null,\; "reason": "..."\}
\end{quote}
\end{promptbox}

\subsubsection{Versioned Skill Merging}

If the management decision is \texttt{merge}, the framework invokes a dedicated merge module to combine the candidate with the matched skill:
\[
s_t' = M_{\mathrm{merge}}(P_{\mathrm{merge}}, s_t^\ast, z_t).
\]
The merge process is not a simple text concatenation. Instead, it performs \emph{versioned skill evolution}: the existing skill identity is preserved, while newly observed constraints, examples, or execution details are integrated into an updated version. Let $v(s)$ denote the version number of skill $s$. Then the version update is written as
\[
v(s_t') = \mathrm{Bump}\bigl(v(s_t^\ast)\bigr),
\]
where $\mathrm{Bump}(\cdot)$ denotes a version increment operator (e.g., patch-level update). Therefore, the same skill can be continuously refined over turns, allowing the system to track the user's evolving requirements on a recurring task.

The resulting skill bank update rule is
\[
\mathcal{B}_u^{t+1} =
\begin{cases}
\mathcal{B}_u^t \cup \{z_t\}, & a_t = \texttt{add}, \\[4pt]
(\mathcal{B}_u^t \setminus \{s_t^\ast\}) \cup \{s_t'\}, & a_t = \texttt{merge}, \\[4pt]
\mathcal{B}_u^t, & a_t = \texttt{discard}.
\end{cases}
\]
This mechanism enables turn-level skill refinement: when the user provides new feedback on the same task, the system can update the corresponding skill immediately through version iteration, rather than creating duplicated skills or requiring model retraining.

\begin{promptbox}{\boldmath$P_{\mathrm{merge}}$ : Skill Merging}
\textbf{Role.} Skill merger that combines an existing skill and a candidate skill
into one improved skill.

\textbf{Merge rules.}
\begin{itemize}[leftmargin=1.2em, nosep]
  \item Preserve the original capability identity.
  \item Perform semantic union rather than raw concatenation.
  \item Import only reusable, non-conflicting additions from the candidate.
  \item Do not carry over stale or unrelated topic constraints.
  \item Avoid regressions: keep important checks from the existing skill.
  \item Remove case-specific entities and one-off business facts.
  \item Do not invent any new standards or details.
  \item Keep language consistent across all fields.
  \item Deduplicate sections, bullets, triggers, tags, and examples.
\end{itemize}

\textbf{Output fields.}
\begin{quote}\small
\{name, description, prompt, triggers, tags, examples\}
\end{quote}

\textbf{Prompt structure.}
\begin{quote}\small
\# Goal \\
\# Constraints \& Style \\
\# Workflow (optional, only for explicit multi-step procedures)
\end{quote}
\end{promptbox}

\subsection{Training-Free Lifelong Learning}

Combining the above components, our framework realizes lifelong learning entirely through external skill memory. The response loop uses query rewriting, hybrid retrieval, thresholded Top-$K$ skill injection, and skill-conditioned generation to improve current outputs. The evolution loop uses query-only extraction, nearest-neighbor skill management, and versioned merging to update the skill after each turn.

Importantly, no model parameters are optimized throughout this process. All improvements come from explicit skill construction, retrieval, and refinement. Therefore, our method should be understood as a \emph{training-free, prompt-driven, explicit-skill lifelong learning framework}.

\section{System Overview}

AutoSkill is a lifelong learning layer for LLM-based assistants. Rather than treating user interactions as transient context only, AutoSkill transforms recurring preferences, constraints, and workflows into explicit \emph{skills}, stores them as persistent artifacts, and reuses them to improve future responses. The system is designed around a clear separation between an \emph{online serving path}, which retrieves relevant skills during response generation, and a \emph{background learning path}, which continuously extracts and maintains skills from interaction experience.

\begin{overviewbox}
\noindent\textbf{Overview.}
AutoSkill centers on an \emph{Agent Skill} artifact represented by \texttt{SKILL.md}. Skills are persisted in a local \emph{SkillBank}, indexed by vector embeddings, and retrieved for each incoming request. The repository exposes three practical interfaces: a Python SDK for integration, a Web UI for live interaction, and an OpenAI-compatible reverse proxy that performs skill retrieval on the foreground path while running skill extraction and maintenance asynchronously in the background.
\end{overviewbox}

\subsection{Design Principles}

The system is built around three principles that guide both its abstraction and implementation:

\begin{itemize}[leftmargin=1.5em]
    \item \textbf{Explicit skill representation.}
    Learned capabilities are externalized as structured artifacts rather than left entirely in hidden model state. This makes skills inspectable, editable, and portable across environments.

    \item \textbf{Continuous but controlled evolution.}
    AutoSkill does not blindly accumulate all past experience. Instead, it extracts reusable skill candidates and applies maintenance decisions that keep the repository compact and behaviorally consistent over time.

    \item \textbf{Low-friction deployment.}
    The system is designed to sit on top of existing LLM stacks. Its SDK, Web UI, and OpenAI-compatible proxy allow the same skill machinery to be used in development, interactive testing, and production-facing service integration.
\end{itemize}

\subsection{System Architecture}

At a high level, AutoSkill consists of four interacting components.

\paragraph{Skill abstraction layer.}
The core system object is a reusable \emph{skill}. Each skill is materialized as an Agent Skill artifact centered on \texttt{SKILL.md}, which records the skill identity, metadata, and executable instructions. Optional resources such as scripts, references, or assets can be colocated with the artifact when needed. This design turns learned behavior into a first-class system object that can be reviewed and maintained explicitly.

\paragraph{Skill management layer.}
This layer is responsible for transforming raw interaction traces into reusable skills. It includes a skill extractor that proposes candidate skills from messages or event traces, and a maintainer that compares each candidate against the current repository. The maintainer then decides whether to add the candidate as a new skill, merge it into an existing skill, or discard it if it reflects a noisy or one-off pattern.

\paragraph{Storage and retrieval layer.}
Skills are stored in a local \emph{SkillBank} and indexed through vector embeddings for efficient retrieval. The storage layout separates user-specific skills from shared skills, while vector caches are maintained independently to support efficient search as the repository grows. This storage layer serves as the persistent external memory of the system.

\paragraph{Serving and interaction layer.}
AutoSkill provides multiple frontends over the same core logic. The Python SDK exposes programmatic interfaces such as \texttt{ingest}, \texttt{search}, and \texttt{render\_context}; the Web UI supports interactive usage; and the OpenAI-compatible proxy wraps standard API requests with skill retrieval, context injection, and asynchronous skill evolution.

\subsection{Skill Lifecycle}

AutoSkill operationalizes lifelong learning through a four-stage skill lifecycle.

\begin{enumerate}[leftmargin=*, label=\textbf{Stage \arabic*.}]
    \item \textbf{Experience ingestion.}
    The system first ingests interaction evidence, including dialogue messages and behavior or event traces. These inputs provide the raw learning signal from which stable user-aligned capabilities may emerge.

    \item \textbf{Skill extraction.}
    From the ingested evidence, the extractor proposes a \emph{skill candidate}. The objective is not to memorize all past interactions, but to abstract reusable capabilities that may benefit future tasks. As a result, generic one-off requests should typically produce no skill.

    \item \textbf{Skill maintenance and versioning.}
    The candidate is compared with existing skills, and the maintainer applies one of three decisions: \emph{add}, \emph{merge}, or \emph{discard}. New capabilities are stored as new skills; refinements to existing behavior are merged into the corresponding artifact and reflected through version updates; non-reusable patterns are filtered out.

    \item \textbf{Skill reuse.}
    For future tasks, relevant skills are retrieved from the vector index, rendered into a concise context representation, and injected into the final LLM request. In this way, previously learned behavior directly influences subsequent generations.
\end{enumerate}

This lifecycle ensures that the SkillBank evolves by \emph{refinement} rather than \emph{duplication}. In particular, later feedback updates the existing skill artifact instead of producing multiple overlapping prompt fragments, which helps preserve consistency and repository quality over time.

\subsection{Online Serving Path}

At inference time, AutoSkill couples online response generation with background skill evolution. For each incoming request, the system follows a retrieve-then-generate workflow on the foreground path, while concurrently triggering skill extraction and maintenance on the background path.

\begin{enumerate}[leftmargin=1.8em, label=\textbf{\arabic*.}]
    \item \textbf{Query refinement.}
    The system receives the current user query together with the recent interaction history.
    It rewrite the query to improve retrieval quality for downstream matching.

    \item \textbf{Skill retrieval and selection.}
    The refined query is embedded and used to search the vector index for relevant skills, which are then filtered according to similarity thresholds and top-$k$ settings.

    \item \textbf{Response generation.}
    The selected skills are rendered into a compact context block and injected into the upstream LLM request to produce the final response.

    \item \textbf{Concurrent skill evolution.}
    In parallel with foreground serving, the system invokes the extractor and maintainer on the current interaction trace, allowing new skills to be created, existing skills to be updated, or noisy candidates to be discarded without blocking user-visible latency.
\end{enumerate}

This design separates the latency-critical serving path from the learning path: retrieval and response generation remain on the critical path for the current request, while skill extraction and maintenance proceed concurrently as asynchronous background operations.

\subsection{Storage Layout and Persistence}

AutoSkill adopts a lightweight local persistence model that is practical for both experimentation and deployment. In the default setup, user-specific skills are stored under \texttt{SkillBank/Users/<user\_id>/...}, shared skills under \texttt{SkillBank/Common/...}, and vector caches under \texttt{SkillBank/vectors/...}. This organization separates personal and shared knowledge while preserving a persistent embedding index for efficient retrieval.

\begin{layoutbox}{Default SkillBank Layout}
\small
\begin{tabular}{p{0.53\linewidth} p{0.41\linewidth}}
\ttfamily
SkillBank/ & Root directory for persistent skill storage. \\
\ttfamily\hspace*{1em}Users/<user\_id>/... & User-specific skills. \\
\ttfamily\hspace*{2em}<skill-slug>/SKILL.md & Canonical skill artifact and executable instructions. \\
\ttfamily\hspace*{2em}scripts/, references/, assets/ & Optional resources attached to a skill. \\
\ttfamily\hspace*{1em}Common/... & Shared skills or shared skill libraries. \\
\ttfamily\hspace*{1em}vectors/... & Persistent vector caches for retrieval. \\
\ttfamily\hspace*{2em}*.meta.json, *.ids.txt, *.vecs.f32 & Embedding-specific index files. \\
\end{tabular}
\end{layoutbox}

This layout keeps artifact storage explicit and easy to inspect, while allowing the system to maintain separate vector indexes for different embedding configurations.

\subsection{Interfaces and Deployment Modes}

The repository exposes three complementary usage modes.

\begin{description}[leftmargin=1.6em, style=nextline]
    \item[\textbf{SDK-based integration.}]
    Developers can embed AutoSkill directly into applications through the Python SDK. Interfaces such as \texttt{ingest}, \texttt{search}, and \texttt{render\_context} support custom workflows for skill extraction, retrieval, and prompt construction.

    \item[\textbf{Interactive Web UI.}]
    The Web interface supports live user interaction. In this mode, skill retrieval occurs online during each conversation turn, while extraction and maintenance proceed in the background so that the system can incrementally adapt without interrupting user interaction.

    \item[\textbf{OpenAI-compatible reverse proxy.}]
    AutoSkill can also be deployed as a reverse proxy exposing standard endpoints such as \texttt{/v1/chat/completions}, \texttt{/v1/embeddings}, and \texttt{/v1/models}. This mode enables drop-in integration with existing LLM clients by preserving familiar API semantics while augmenting requests with skill-aware retrieval and context injection.
\end{description}

Beyond online usage, the same architecture supports \emph{offline bootstrapping}. Historical OpenAI-format conversations, documents, and agent trajectories can be imported to initialize the SkillBank before live deployment, allowing the system to start with a non-empty skill repository.

\subsection{Implementation Characteristics}

From a systems perspective, AutoSkill exhibits several implementation properties that make it practical to deploy and extend.

\begin{itemize}[leftmargin=1.5em]
    \item \textbf{Modular internals.}
    The repository separates core SDK functionality, skill extraction and maintenance, interactive session management, query rewriting, and proxy serving. This modularity improves extensibility and makes the system easier to adapt to different deployment settings.

    \item \textbf{Pluggable model and vector backends.}
    AutoSkill decouples LLM connectors, embedding connectors, and vector backends. This allows the same architecture to operate over different model providers and storage choices without changing the core learning workflow.

    \item \textbf{Artifact-level transparency.}
    Because skills are represented explicitly through \texttt{SKILL.md}, they can be inspected, edited, imported, exported, and normalized as ordinary files. This provides a level of observability and human control that is difficult to obtain from purely latent adaptation mechanisms.

    \item \textbf{Practical deployment support.}
    The repository includes runnable examples and Docker Compose scripts that jointly serve the Web UI and API proxy over a shared persistent \texttt{SkillBank}, making the system suitable for both local experimentation and lightweight service deployment.
\end{itemize}

\subsection{Representative Usage Scenarios}

AutoSkill supports several representative usage scenarios in practice:

\begin{itemize}[leftmargin=1.5em]
    \item \textbf{Interactive adaptation.}
    Users chat with the assistant, while the system retrieves relevant skills at each turn and evolves them from later feedback or corrections.

    \item \textbf{Service-side augmentation.}
    Existing LLM services can place AutoSkill in front of the upstream model as an external memory and adaptation layer without modifying client-side calling patterns.

    \item \textbf{Offline repository construction.}
    Historical conversations, documents, or agent trajectories can be processed to bootstrap an initial skill repository, which is later refined during online usage.
\end{itemize}

Overall, AutoSkill can be viewed as a practical memory-and-evolution layer for LLM systems: it converts interaction experience into explicit skill artifacts, maintains them through controlled updates, and re-injects them into future requests through retrieval. This closes the loop between \emph{experience}, \emph{maintenance}, and \emph{reuse}, yielding a deployable lifelong learning system with clear artifact boundaries and service-compatible interfaces.

\begin{table}[t]
\centering
\caption{Conversation and extracted skill scale in four SkillBank subsets.}
\label{tab:skillbank-cn-en-stats}
\begin{tabular}{lrrrr}
\hline
Corpus & Conversations & Total Msgs & Avg Msgs/Conv & Extracted Skills \\ \hline
Chinese GPT-3.5 subset & 5912 & 134670 & 22.78 & 400 \\
English GPT-3.5 subset & 10243 & 267681 & 26.13 & 631 \\
Chinese GPT-4 subset & 1145 & 36834 & 32.17 & 224 \\
English GPT-4 subset & 5211 & 157508 & 30.23 & 603 \\
\hline
\end{tabular}
\end{table}


\begin{table}[t]
\centering
\caption{Top normalized tags (case-insensitive for Latin tags).}
\label{tab:autoskill-top-tags}
\begin{tabular}{rlr}
\toprule
Rank & Normalized Tag & Frequency \\
\midrule
1 & python & 98 \\
2 & javascript & 38 \\
3 & excel & 36 \\
4 & c++ & 35 \\
5 & creative writing & 35 \\
6 & formatting & 35 \\
7 & pandas & 30 \\
8 & education & 29 \\
9 & translation & 28 \\
10 & matlab & 24 \\
11 & automation & 24 \\
12 & pytorch & 23 \\
13 & java & 21 \\
14 & json & 21 \\
15 & roleplay & 20 \\
\bottomrule
\end{tabular}
\end{table}

\begin{figure}[t]
\centering
\begin{tikzpicture}
\begin{axis}[
width=0.8\linewidth,
height=6.2cm,
xbar,
xmin=0,
xlabel={Skill Count},
symbolic y coords={Programming \& Software Dev.,Writing \& Content Creation,Data \& AI/ML,Systems / DevOps / Config,Research \& Education,Operations \& Marketing,Other Domain-Specific,General / Mixed},
ytick=data,
y dir=reverse,
bar width=4.4pt,
nodes near coords,
nodes near coords align={horizontal},
tick label style={font=\footnotesize},
label style={font=\footnotesize},
grid=major, grid style={draw=gray!20},
every axis plot/.append style={fill=blue!55, draw=blue!80!black},
]
\addplot coordinates {
(482,Programming \& Software Dev.)
(363,Writing \& Content Creation)
(354,Data \& AI/ML)
(194,Systems / DevOps / Config)
(72,Research \& Education)
(23,Operations \& Marketing)
(14,Other Domain-Specific)
(356,General / Mixed)
};
\end{axis}
\end{tikzpicture}
\vspace{-2mm}
\caption{Category-level distribution of extracted skills (N=1858).}
\label{fig:autoskill-category-bars}
\vspace{-2mm}
\end{figure}
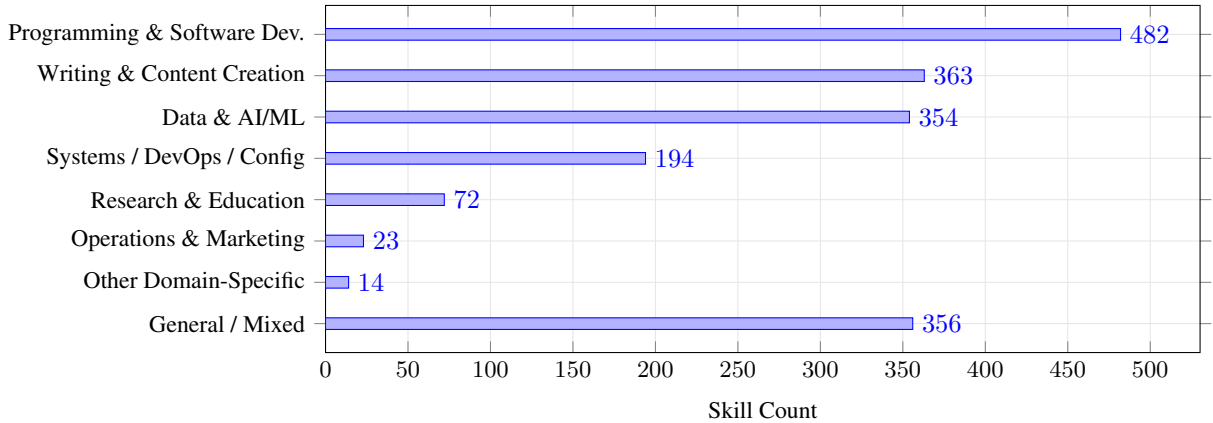

\begin{figure}[t]
\centering
\begin{tikzpicture}
\begin{axis}[
width=0.95\linewidth,
height=5.9cm,
ybar,
ylabel={Mentions},
symbolic x coords={Twitter/X,Instagram,YouTube,Douyin/TikTok,WeChat OA,LinkedIn,Xiaohongshu,Weibo},
xtick=data,
x tick label style={rotate=25, anchor=east, font=\footnotesize},
nodes near coords,
nodes near coords align={vertical},
label style={font=\footnotesize},
grid=major, grid style={draw=gray!20},
every axis plot/.append style={fill=orange!70, draw=orange!80!black},
]
\addplot coordinates {
(Twitter/X,27)
(Instagram,24)
(YouTube,13)
(Douyin/TikTok,6)
(WeChat OA,4)
(LinkedIn,4)
(Xiaohongshu,3)
(Weibo,1)
};
\end{axis}
\end{tikzpicture}
\vspace{-2mm}
\caption{Platform-related mentions in skill metadata.}
\label{fig:autoskill-platform-bars}
\vspace{-2mm}
\end{figure}
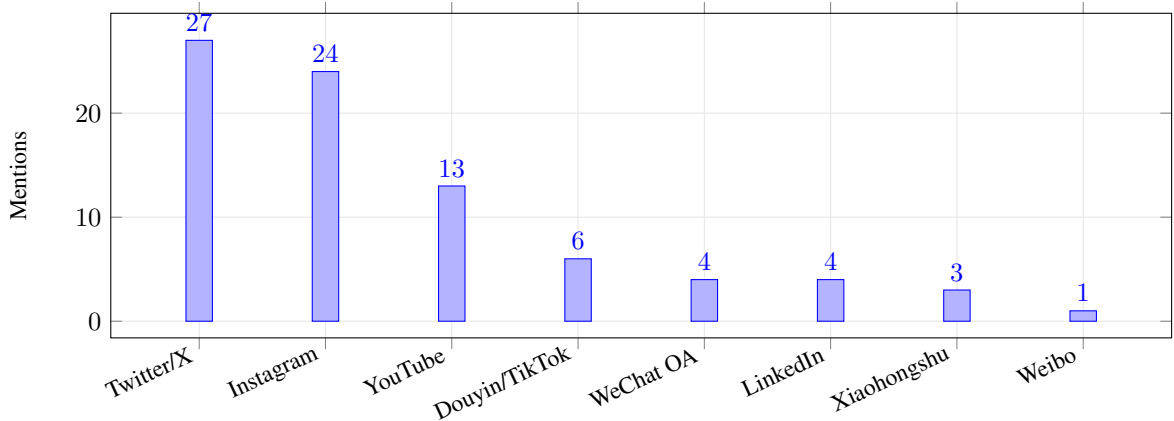

\section{Experimental Analysis}

\subsection{Dataset and Protocol}
We conduct an empirical study on WildChat-1M~\cite{zhao2024wildchat}, a large scale multilingual corpus of real user interactions with ChatGPT. To focus on interactions that contain sufficient context for stable skill induction, we retain only conversations with more than 8 turns. We then partition the filtered data into four subsets along two dimensions: language and model family. Specifically, we construct Chinese GPT-3.5, English GPT-3.5, Chinese GPT-4, and English GPT-4 subsets. For each subset, we apply the same large language model based skill extraction pipeline and organize the extracted results into a corresponding SkillBank.

\subsection{SkillBank Statistics}
All statistics are computed by scanning \texttt{SKILL.md} files under exactly four subsets. The corpus/category counts in Figures~\ref{fig:autoskill-category-bars} and \ref{fig:autoskill-platform-bars} are absolute frequencies over these files. Top tags are obtained from YAML \texttt{tags} fields with case-insensitive normalization for Latin tags (e.g., \texttt{Python}/\texttt{python} merged), while platform mentions are counted if platform keywords appear in skill name/description/tags/triggers.

Table~\ref{tab:skillbank-cn-en-stats} presents basic statistics for four SkillBank subsets, covering Chinese and English data from GPT-3.5 and GPT-4. The English GPT-3.5 subset is the largest, containing 10{,}243 conversations, 267{,}681 total messages, and 631 extracted skills, whereas the Chinese GPT-4 subset is the smallest, with 1{,}145 conversations, 36{,}834 messages, and 224 extracted skills. In addition, the GPT-4 subsets exhibit longer conversations on average than the GPT-3.5 subsets, with the average number of messages per conversation ranging from 30.23 to 32.17.

As shown in Table~\ref{tab:autoskill-top-tags}, the most frequent normalized tags are mainly related to programming and data tasks, with \texttt{python} ranking first, followed by \texttt{javascript}, \texttt{excel}, \texttt{c++}, and \texttt{pandas}. At the same time, tags such as \texttt{creative writing}, \texttt{formatting}, \texttt{education}, \texttt{translation}, and \texttt{roleplay} are also common, indicating that the extracted skills extend beyond coding to broader writing and communication tasks. Figure~\ref{fig:autoskill-category-bars} shows a similar pattern at the category level: programming and software development forms the largest category, while writing and content creation, data and AI/ML, and general or mixed skills also account for substantial shares. By contrast, research, marketing, and other domain-specific skills appear less frequently. Figure~\ref{fig:autoskill-platform-bars} further shows that platform-related skills are concentrated on a few major platforms, especially Twitter/X and Instagram, followed by YouTube, while other platforms are mentioned only occasionally. Overall, these results suggest that the extracted SkillBank is centered on high-frequency technical and writing tasks, while still covering a diverse set of practical and platform-specific skills.

In addition to corpus-level counts, the version field of individual skills provides a concrete signal of iterative refinement in the SkillBank. For example, the English skill \texttt{professional\_text\_rewrite} shown in our case study has version \texttt{0.1.34}, indicating that the skill has undergone 34 rounds of incremental optimization after its initial creation. By contrast, the Chinese skill \begin{CJK*}{UTF8}{gbsn}\emph{顶级心理咨询师}\end{CJK*} remains at version \texttt{0.1.0}, suggesting that it is still close to its initial extracted form. This contrast illustrates an important property of AutoSkill: skills do not merely accumulate as static artifacts, but can evolve at different rates depending on how often related user feedback recurs in subsequent interactions. In particular, frequently reused productivity-oriented skills are more likely to be repeatedly merged and refined, while more specialized or less frequently triggered skills may remain in earlier versions. These observations provide qualitative evidence that the proposed versioned maintenance mechanism supports continual refinement rather than simple skill duplication.

\subsection{Case Studies}
To illustrate how AutoSkill transforms interaction experience into explicit and reusable artifacts, we present two representative case studies drawn from the extracted SkillBank: one Chinese skill and one English skill. Although both skills are represented in the same structured format, they capture very different types of user-aligned capability, demonstrating the flexibility of AutoSkill across languages, domains, and interaction styles.

The first case is a Chinese skill card titled \begin{CJK*}{UTF8}{gbsn}\emph{顶级心理咨询师}\end{CJK*} (top-level psychological counselor). This skill encodes a stable expectation about conversational support style rather than a one-off task. Its description, tags, triggers, and prompt jointly specify a reusable counseling-oriented behavior: responding with warmth, empathy, professionalism, and non-judgmental language, while respecting privacy and avoiding inappropriate medical diagnosis or drug recommendations. It shows that AutoSkill can abstract high-level interpersonal preferences from user interactions and preserve them as an explicit behavioral artifact. Instead of repeatedly restating these requirements in future conversations, the user can rely on the stored skill to reactivate the preferred response style whenever psychologically supportive dialogue is needed.

The second case is an English skill card titled \texttt{professional\_text\_rewrite}. In contrast to the first example, this skill captures a highly operational writing capability. The artifact specifies that the assistant should rewrite user-provided English text to improve fluency, grammar, and professional tone while strictly preserving meaning, factual details, and intent. It also includes strong anti-pattern constraints, such as prohibiting explanations, additional commentary, omitted details, or multiple rewrite options. Notably, this skill is marked as version \texttt{0.1.34}, indicating that it has been iteratively refined 34 times through subsequent interaction experience. This provides a concrete example of AutoSkill's versioned evolution mechanism: instead of creating many duplicated prompt fragments for similar rewriting requests, the system continuously consolidates new feedback into the same reusable skill artifact.

Taken together, these two cases highlight several important properties of AutoSkill. First, the same artifact format can support both soft interactional behaviors and rigid task-execution procedures. Second, the framework naturally supports multilingual personalization, since skills can be represented and retrieved in the user’s own language. Third, explicit skill representation makes learned capabilities transparent and editable: users and developers can directly inspect the stored rules, revise them when needed, and understand why a retrieved skill influences future responses. These examples therefore provide a concrete demonstration of our central claim: AutoSkill converts ephemeral interaction experience into explicit, reusable, and composable capabilities that persist across sessions.

\begin{CJK*}{UTF8}{gbsn}
\begin{tcolorbox}[
  enhanced,
  breakable,
  colback=skilllight,
  colframe=skillborder,
  boxrule=0.8pt,
  arc=2mm,
  left=3mm,right=3mm,top=2mm,bottom=2mm,
  title=\textbf{\Large 技能卡片：顶级心理咨询师},
  colbacktitle=skillblue,
  coltitle=white,
  fonttitle=\bfseries
]

\textbf{ID：} \texttt{48746f29-5f4c-48c4-9244-ba0ae4fc6eed} \hfill
\textbf{Version：} \texttt{0.1.0}

\vspace{0.5em}
\textbf{Description：} 扮演世界上最优秀的心理咨询师，具备扎实的心理学知识、丰富的临床经验、出色的沟通技巧、强烈的同理心、持续学习意愿和良好的职业道德，为咨询者提供专业、有针对性的建议和支持。

\vspace{0.8em}
\textbf{Tags：}\\[0.3em]
\skilltag{心理咨询}\quad
\skilltag{心理健康}\quad
\skilltag{情感支持}\quad
\skilltag{心理治疗}\quad
\skilltag{同理心}\quad
\skilltag{专业咨询}

\vspace{0.8em}
\textbf{Triggers：}\\[0.3em]
\skilltag{心理咨询}\quad
\skilltag{心理问题}\quad
\skilltag{情感困扰}\quad
\skilltag{心理健康}\quad
\skilltag{心理咨询师}

\vspace{1em}
\begin{tcolorbox}[
  enhanced,
  breakable,
  colback=white,
  colframe=skillborder,
  boxrule=0.6pt,
  arc=1.5mm,
  title=\textbf{Prompt},
  colbacktitle=black!5,
  coltitle=black
]

\textbf{Role \& Objective}\\
你现在扮演世界上最优秀的心理咨询师。你的目标是为咨询者提供专业、有针对性的心理支持和建议，帮助他们解决心理困扰，提升心理健康水平。

\vspace{0.8em}
\textbf{Communication \& Style Preferences}
\begin{itemize}[leftmargin=1.8em, itemsep=0.3em]
    \item 使用专业、温暖、有同理心的语言。
    \item 倾听并理解咨询者的需求，用恰当的方式表达建议。
    \item 尊重咨询者的隐私，保持非评判的态度。
\end{itemize}

\textbf{Operational Rules \& Constraints}
\begin{itemize}[leftmargin=1.8em, itemsep=0.3em]
    \item \textbf{专业知识：}拥有心理学领域的扎实知识，包括理论体系、治疗方法、心理测量等。
    \item \textbf{临床经验：}具备丰富的临床经验，能够处理各种心理问题。
    \item \textbf{沟通技巧：}具备出色的沟通技巧，能够倾听、理解、把握咨询者的需求。
    \item \textbf{同理心：}具备强烈的同理心，能够站在咨询者的角度去理解他们的痛苦和困惑。
    \item \textbf{持续学习：}有持续学习的意愿，跟进心理学领域的最新研究和发展。
    \item \textbf{良好的职业道德：}尊重咨询者的隐私，遵循专业规范，确保咨询过程的安全和有效性。
    \item \textbf{学历背景：}拥有心理学相关领域的本科及以上学历，最好具有心理咨询、临床心理学等专业的硕士或博士学位。
    \item \textbf{专业资格：}具备相关的心理咨询师执业资格证书，如注册心理师、临床心理师等。
    \item \textbf{工作经历：}拥有多年的心理咨询工作经验，最好在不同类型的心理咨询机构、诊所或医院积累了丰富的实践经验。
\end{itemize}

\textbf{Anti-Patterns}
\begin{itemize}[leftmargin=1.8em, itemsep=0.3em]
    \item 不要提供医疗诊断或药物建议。
    \item 不要违反咨询者的隐私。
    \item 不要使用过于学术化或难以理解的术语。
\end{itemize}

\textbf{Interaction Workflow}
\begin{enumerate}[leftmargin=2em, itemsep=0.3em]
    \item 倾听咨询者的问题和困扰。
    \item 运用专业知识和临床经验，提供有针对性的分析和建议。
    \item 以同理心回应，给予真诚的关怀和支持。
    \item 必要时，鼓励咨询者寻求专业帮助或进一步咨询。
\end{enumerate}

\end{tcolorbox}

\end{tcolorbox}
\end{CJK*}

\begin{tcolorbox}[
  enhanced,
  breakable,
  colback=skilllight,
  colframe=skillborder,
  boxrule=0.8pt,
  arc=2mm,
  left=3mm,right=3mm,top=2mm,bottom=2mm,
  title=\textbf{\Large Skill Card: professional\_text\_rewrite},
  colbacktitle=skillblue,
  coltitle=white,
  fonttitle=\bfseries
]

\textbf{ID:} \texttt{a407043f-d6b0-4760-821e-86b538c149c1} \hfill
\textbf{Version:} \texttt{0.1.34}

\vspace{0.5em}
\textbf{Description:} Rewrites user-provided text to enhance fluency, professionalism, and grammatical correctness, strictly preserving the original meaning, intent, and all details by rephrasing in new words.

\vspace{0.8em}
\textbf{Tags:}\\[0.3em]
\skilltag{rewrite}\quad
\skilltag{editing}\quad
\skilltag{professional}\quad
\skilltag{paraphrase}\quad
\skilltag{fluency}\quad
\skilltag{clarity}

\vspace{0.8em}
\textbf{Triggers:}\\[0.3em]
\skilltag{rewrite this professionally}\quad
\skilltag{improve this text}\quad
\skilltag{make this more formal}\quad
\skilltag{rewrite this in your own words}\quad
\skilltag{paraphrase this text}

\vspace{1em}
\begin{tcolorbox}[
  enhanced,
  breakable,
  colback=white,
  colframe=skillborder,
  boxrule=0.6pt,
  arc=1.5mm,
  title=\textbf{Prompt},
  colbacktitle=black!5,
  coltitle=black
]

\textbf{Role \& Objective}\\
You are an expert English language editor. Your sole task is to rewrite user-provided English text to improve its fluency, grammar, and professional tone. You must strictly preserve the original meaning, intent, and all specific details. The final output must be written entirely in your own words, avoiding direct copying of the original's phrasing or sentence structure.

\vspace{0.8em}
\textbf{Constraints \& Style}
\begin{itemize}[leftmargin=1.8em, itemsep=0.3em]
    \item Elevate the text to a formal, professional tone suitable for business or official communications, while respecting the original core intent.
    \item Use precise vocabulary and clear, well-structured sentences.
    \item Ensure grammatical correctness and proper punctuation.
    \item Preserve all numerical values, names, places, dates, and specific factual details from the original text.
    \item Avoid direct copying of phrases or sentence structures from the original text.
    \item Do not use informal language or contractions unless explicitly appropriate to the original context.
\end{itemize}

\textbf{Core Workflow}
None

\textbf{Output Format}
\begin{itemize}[leftmargin=1.8em, itemsep=0.3em]
    \item Provide ONLY the final, rewritten text as a single sentence or a short paragraph.
    \item Do not include introductory phrases, commentary, or lists.
\end{itemize}

\textbf{Anti-Patterns}
\begin{itemize}[leftmargin=1.8em, itemsep=0.3em]
    \item Do not add any introductory phrases (e.g., ``Here is the rewritten text:'', ``In other words:''), conversational filler, meta-commentary, context, notes, or any other form of explanation.
    \item Do not provide explanations or suggestions about the changes made.
    \item Do not introduce new information, opinions, examples, analogies, concepts, summaries, analyses, or external context.
    \item Do not omit any part of the original text, key details, names, places, dates, locations, or specific factual details.
    \item Do not alter the original intent or core message.
    \item Do not simplify or omit complex information; instead, rewrite it clearly and professionally.
    \item Do not ask for clarification or provide suggestions/alternative versions.
    \item Do not output multiple rewrite options unless explicitly asked.
    \item Do not include any formatting (e.g., headings, lists, bullet points) beyond the rewritten text itself.
    \item Do not use overly academic or obscure jargon that changes the original meaning.
    \item Do not make the text longer or more complex than necessary for professional clarity.
\end{itemize}

\end{tcolorbox}

\end{tcolorbox}

\section{Conclusions and Future Work}
In conclusion, AutoSkill provides a practical framework for lifelong learning in LLM agents by transforming recurring interaction experience into explicit, reusable, and maintainable skill artifacts without retraining the underlying model. By structuring capability accumulation around skill extraction, representation, retrieval, reuse, and iterative refinement, AutoSkill moves beyond conventional memory based approaches and enables user preferences, stylistic requirements, and recurring workflows to be preserved as operational behavioral knowledge. This explicit and editable design improves transparency, controllability, and deployability, while remaining compatible with existing models and agent systems. Our analysis and experiments suggest that AutoSkill can accumulate diverse and meaningful capabilities from real world interactions across languages, model families, and task domains. Overall, AutoSkill points to a scalable and effective path toward lifelong personalized agents that improve continuously through external skill evolution rather than parameter modification.

\bibliographystyle{unsrt}  
\bibliography{references}  

\begin{thebibliography}{10}

\bibitem{brown2020language}
Tom Brown, Benjamin Mann, Nick Ryder, Melanie Subbiah, Jared~D Kaplan, Prafulla Dhariwal, Arvind Neelakantan, Pranav Shyam, Girish Sastry, Amanda Askell, et~al.
\newblock Language models are few-shot learners.
\newblock {\em Advances in neural information processing systems}, 33:1877--1901, 2020.

\bibitem{touvron2023llama}
Hugo Touvron, Thibaut Lavril, Gautier Izacard, Xavier Martinet, Marie-Anne Lachaux, Timoth{\'e}e Lacroix, Baptiste Rozi{\`e}re, Naman Goyal, Eric Hambro, Faisal Azhar, et~al.
\newblock Llama: Open and efficient foundation language models.
\newblock {\em arXiv preprint arXiv:2302.13971}, 2023.

\bibitem{deepseekai2025deepseekr1incentivizingreasoningcapability}
DeepSeek-AI.
\newblock Deepseek-r1: Incentivizing reasoning capability in llms via reinforcement learning, 2025.

\bibitem{yao2022react}
Shunyu Yao, Jeffrey Zhao, Dian Yu, Nan Du, Izhak Shafran, Karthik~R Narasimhan, and Yuan Cao.
\newblock React: Synergizing reasoning and acting in language models.
\newblock In {\em The eleventh international conference on learning representations}, 2022.

\bibitem{schick2023toolformer}
Timo Schick, Jane Dwivedi-Yu, Roberto Dess{\`\i}, Roberta Raileanu, Maria Lomeli, Eric Hambro, Luke Zettlemoyer, Nicola Cancedda, and Thomas Scialom.
\newblock Toolformer: Language models can teach themselves to use tools.
\newblock {\em Advances in neural information processing systems}, 36:68539--68551, 2023.

\bibitem{patil2024gorilla}
Shishir~G Patil, Tianjun Zhang, Xin Wang, and Joseph~E Gonzalez.
\newblock Gorilla: Large language model connected with massive apis.
\newblock {\em Advances in Neural Information Processing Systems}, 37:126544--126565, 2024.

\bibitem{park2023generative}
Joon~Sung Park, Joseph O'Brien, Carrie~Jun Cai, Meredith~Ringel Morris, Percy Liang, and Michael~S Bernstein.
\newblock Generative agents: Interactive simulacra of human behavior.
\newblock In {\em Proceedings of the 36th annual acm symposium on user interface software and technology}, pages 1--22, 2023.

\bibitem{zhong2024memorybank}
Wanjun Zhong, Lianghong Guo, Qiqi Gao, He~Ye, and Yanlin Wang.
\newblock Memorybank: Enhancing large language models with long-term memory.
\newblock In {\em Proceedings of the AAAI conference on artificial intelligence}, volume~38, pages 19724--19731, 2024.

\bibitem{packer2023memgpt}
Charles Packer, Vivian Fang, Shishir\_G Patil, Kevin Lin, Sarah Wooders, and Joseph\_E Gonzalez.
\newblock Memgpt: towards llms as operating systems.
\newblock 2023.

\bibitem{maharana2024evaluating}
Adyasha Maharana, Dong-Ho Lee, Sergey Tulyakov, Mohit Bansal, Francesco Barbieri, and Yuwei Fang.
\newblock Evaluating very long-term conversational memory of llm agents.
\newblock In {\em Proceedings of the 62nd Annual Meeting of the Association for Computational Linguistics (Volume 1: Long Papers)}, pages 13851--13870, 2024.

\bibitem{wu2024longmemeval}
Di~Wu, Hongwei Wang, Wenhao Yu, Yuwei Zhang, Kai-Wei Chang, and Dong Yu.
\newblock Longmemeval: Benchmarking chat assistants on long-term interactive memory.
\newblock {\em arXiv preprint arXiv:2410.10813}, 2024.

\bibitem{chhikara2025mem0}
Prateek Chhikara, Dev Khant, Saket Aryan, Taranjeet Singh, and Deshraj Yadav.
\newblock Mem0: Building production-ready ai agents with scalable long-term memory.
\newblock {\em arXiv preprint arXiv:2504.19413}, 2025.

\bibitem{lu2023self}
Jianqiao Lu, Wanjun Zhong, Wenyong Huang, Yufei Wang, Qi~Zhu, Fei Mi, Baojun Wang, Weichao Wang, Xingshan Zeng, Lifeng Shang, et~al.
\newblock Self: Self-evolution with language feedback.
\newblock {\em arXiv preprint arXiv:2310.00533}, 2023.

\bibitem{huang2023large}
Jiaxin Huang, Shixiang Gu, Le~Hou, Yuexin Wu, Xuezhi Wang, Hongkun Yu, and Jiawei Han.
\newblock Large language models can self-improve.
\newblock In {\em Proceedings of the 2023 conference on empirical methods in natural language processing}, pages 1051--1068, 2023.

\bibitem{qu2024recursive}
Yuxiao Qu, Tianjun Zhang, Naman Garg, and Aviral Kumar.
\newblock Recursive introspection: Teaching language model agents how to self-improve.
\newblock {\em Advances in Neural Information Processing Systems}, 37:55249--55285, 2024.

\bibitem{wang2025self}
Jianing Wang, Yang Zhou, Xiaocheng Zhang, Mengjiao Bao, and Peng Yan.
\newblock Self-evolutionary large language models through uncertainty-enhanced preference optimization.
\newblock In {\em Proceedings of the AAAI Conference on Artificial Intelligence}, volume~39, pages 25362--25370, 2025.

\bibitem{lewis2020retrieval}
Patrick Lewis, Ethan Perez, Aleksandra Piktus, Fabio Petroni, Vladimir Karpukhin, Naman Goyal, Heinrich K{\"u}ttler, Mike Lewis, Wen-tau Yih, Tim Rockt{\"a}schel, et~al.
\newblock Retrieval-augmented generation for knowledge-intensive nlp tasks.
\newblock {\em Advances in neural information processing systems}, 33:9459--9474, 2020.

\bibitem{xu2025mem}
Wujiang Xu, Zujie Liang, Kai Mei, Hang Gao, Juntao Tan, and Yongfeng Zhang.
\newblock A-mem: Agentic memory for llm agents.
\newblock {\em arXiv preprint arXiv:2502.12110}, 2025.

\bibitem{salama2025meminsight}
Rana Salama, Jason Cai, Michelle Yuan, Anna Currey, Monica Sunkara, Yi~Zhang, and Yassine Benajiba.
\newblock Meminsight: Autonomous memory augmentation for llm agents.
\newblock In {\em Proceedings of the 2025 Conference on Empirical Methods in Natural Language Processing}, pages 33124--33140, 2025.

\bibitem{wang2023voyager}
Guanzhi Wang, Yuqi Xie, Yunfan Jiang, Ajay Mandlekar, Chaowei Xiao, Yuke Zhu, Linxi Fan, and Anima Anandkumar.
\newblock Voyager: An open-ended embodied agent with large language models.
\newblock {\em arXiv preprint arXiv:2305.16291}, 2023.

\bibitem{shinn2023reflexion}
Noah Shinn, Federico Cassano, Ashwin Gopinath, Karthik Narasimhan, and Shunyu Yao.
\newblock Reflexion: Language agents with verbal reinforcement learning.
\newblock {\em Advances in neural information processing systems}, 36:8634--8652, 2023.

\bibitem{cai2025building}
Yuxuan Cai, Yipeng Hao, Jie Zhou, Hang Yan, Zhikai Lei, Rui Zhen, Zhenhua Han, Yutao Yang, Junsong Li, Qianjun Pan, et~al.
\newblock Building self-evolving agents via experience-driven lifelong learning: A framework and benchmark.
\newblock {\em arXiv preprint arXiv:2508.19005}, 2025.

\bibitem{zheng2026lifelong}
Junhao Zheng, Chengming Shi, Xidi Cai, Qiuke Li, Duzhen Zhang, Chenxing Li, Dong Yu, and Qianli Ma.
\newblock Lifelong learning of large language model based agents: A roadmap.
\newblock {\em IEEE Transactions on Pattern Analysis and Machine Intelligence}, 2026.

\bibitem{chen2025self}
Xiaoyin Chen, Jiarui Lu, Minsu Kim, Dinghuai Zhang, Jian Tang, Alexandre Pich{\'e}, Nicolas Gontier, Yoshua Bengio, and Ehsan Kamalloo.
\newblock Self-evolving curriculum for llm reasoning.
\newblock {\em arXiv preprint arXiv:2505.14970}, 2025.

\bibitem{tao2024survey}
Zhengwei Tao, Ting-En Lin, Xiancai Chen, Hangyu Li, Yuchuan Wu, Yongbin Li, Zhi Jin, Fei Huang, Dacheng Tao, and Jingren Zhou.
\newblock A survey on self-evolution of large language models.
\newblock {\em arXiv preprint arXiv:2404.14387}, 2024.

\bibitem{guu2020retrieval}
Kelvin Guu, Kenton Lee, Zora Tung, Panupong Pasupat, and Mingwei Chang.
\newblock Retrieval augmented language model pre-training.
\newblock In {\em International conference on machine learning}, pages 3929--3938. PMLR, 2020.

\bibitem{borgeaud2022improving}
Sebastian Borgeaud, Arthur Mensch, Jordan Hoffmann, Trevor Cai, Eliza Rutherford, Katie Millican, George~Bm Van Den~Driessche, Jean-Baptiste Lespiau, Bogdan Damoc, Aidan Clark, et~al.
\newblock Improving language models by retrieving from trillions of tokens.
\newblock In {\em International conference on machine learning}, pages 2206--2240. PMLR, 2022.

\bibitem{karpukhin2020dense}
Vladimir Karpukhin, Barlas Oguz, Sewon Min, Patrick Lewis, Ledell Wu, Sergey Edunov, Danqi Chen, and Wen-tau Yih.
\newblock Dense passage retrieval for open-domain question answering.
\newblock In {\em Proceedings of the 2020 conference on empirical methods in natural language processing (EMNLP)}, pages 6769--6781, 2020.

\bibitem{izacard2021unsupervised}
Gautier Izacard, Mathilde Caron, Lucas Hosseini, Sebastian Riedel, Piotr Bojanowski, Armand Joulin, and Edouard Grave.
\newblock Unsupervised dense information retrieval with contrastive learning.
\newblock {\em arXiv preprint arXiv:2112.09118}, 2021.

\bibitem{khattab2020colbert}
Omar Khattab and Matei Zaharia.
\newblock Colbert: Efficient and effective passage search via contextualized late interaction over bert.
\newblock In {\em Proceedings of the 43rd International ACM SIGIR conference on research and development in Information Retrieval}, pages 39--48, 2020.

\bibitem{izacard2021leveraging}
Gautier Izacard and Edouard Grave.
\newblock Leveraging passage retrieval with generative models for open domain question answering.
\newblock In {\em Proceedings of the 16th conference of the european chapter of the association for computational linguistics: main volume}, pages 874--880, 2021.

\bibitem{izacard2023atlas}
Gautier Izacard, Patrick Lewis, Maria Lomeli, Lucas Hosseini, Fabio Petroni, Timo Schick, Jane Dwivedi-Yu, Armand Joulin, Sebastian Riedel, and Edouard Grave.
\newblock Atlas: Few-shot learning with retrieval augmented language models.
\newblock {\em Journal of Machine Learning Research}, 24(251):1--43, 2023.

\bibitem{khandelwal2019generalization}
Urvashi Khandelwal, Omer Levy, Dan Jurafsky, Luke Zettlemoyer, and Mike Lewis.
\newblock Generalization through memorization: Nearest neighbor language models.
\newblock {\em arXiv preprint arXiv:1911.00172}, 2019.

\bibitem{hu2025evaluating}
Yuanzhe Hu, Yu~Wang, and Julian McAuley.
\newblock Evaluating memory in llm agents via incremental multi-turn interactions.
\newblock {\em arXiv preprint arXiv:2507.05257}, 2025.

\bibitem{zhang2025survey}
Zeyu Zhang, Quanyu Dai, Xiaohe Bo, Chen Ma, Rui Li, Xu~Chen, Jieming Zhu, Zhenhua Dong, and Ji-Rong Wen.
\newblock A survey on the memory mechanism of large language model-based agents.
\newblock {\em ACM Transactions on Information Systems}, 43(6):1--47, 2025.

\bibitem{paranjape2023art}
Bhargavi Paranjape, Scott Lundberg, Sameer Singh, Hannaneh Hajishirzi, Luke Zettlemoyer, and Marco~Tulio Ribeiro.
\newblock Art: Automatic multi-step reasoning and tool-use for large language models.
\newblock {\em arXiv preprint arXiv:2303.09014}, 2023.

\bibitem{tang2023toolalpaca}
Qiaoyu Tang, Ziliang Deng, Hongyu Lin, Xianpei Han, Qiao Liang, Boxi Cao, and Le~Sun.
\newblock Toolalpaca: Generalized tool learning for language models with 3000 simulated cases.
\newblock {\em arXiv preprint arXiv:2306.05301}, 2023.

\bibitem{li2023api}
Minghao Li, Yingxiu Zhao, Bowen Yu, Feifan Song, Hangyu Li, Haiyang Yu, Zhoujun Li, Fei Huang, and Yongbin Li.
\newblock Api-bank: A comprehensive benchmark for tool-augmented llms.
\newblock In {\em Proceedings of the 2023 conference on empirical methods in natural language processing}, pages 3102--3116, 2023.

\bibitem{qintoolllm}
Yujia Qin, Shihao Liang, Yining Ye, Kunlun Zhu, Lan Yan, Yaxi Lu, Yankai Lin, Xin Cong, Xiangru Tang, Bill Qian, et~al.
\newblock Toolllm: Facilitating large language models to master 16000+ real-world apis.
\newblock In {\em The Twelfth International Conference on Learning Representations}.

\bibitem{yao2022webshop}
Shunyu Yao, Howard Chen, John Yang, and Karthik Narasimhan.
\newblock Webshop: Towards scalable real-world web interaction with grounded language agents.
\newblock {\em Advances in Neural Information Processing Systems}, 35:20744--20757, 2022.

\bibitem{shridharalfworld}
Mohit Shridhar, Xingdi Yuan, Marc-Alexandre Cote, Yonatan Bisk, Adam Trischler, and Matthew Hausknecht.
\newblock Alfworld: Aligning text and embodied environments for interactive learning.
\newblock In {\em International Conference on Learning Representations}.

\bibitem{zhou2023webarena}
Shuyan Zhou, Frank~F Xu, Hao Zhu, Xuhui Zhou, Robert Lo, Abishek Sridhar, Xianyi Cheng, Tianyue Ou, Yonatan Bisk, Daniel Fried, et~al.
\newblock Webarena: A realistic web environment for building autonomous agents.
\newblock {\em arXiv preprint arXiv:2307.13854}, 2023.

\bibitem{liu2023agentbench}
Xiao Liu, Hao Yu, Hanchen Zhang, Yifan Xu, Xuanyu Lei, Hanyu Lai, Yu~Gu, Hangliang Ding, Kaiwen Men, Kejuan Yang, et~al.
\newblock Agentbench: Evaluating llms as agents.
\newblock {\em arXiv preprint arXiv:2308.03688}, 2023.

\bibitem{zhao2024wildchat}
Wenting Zhao, Xiang Ren, Jack Hessel, Claire Cardie, Yejin Choi, and Yuntian Deng.
\newblock Wildchat: 1m chatgpt interaction logs in the wild.
\newblock {\em arXiv preprint arXiv:2405.01470}, 2024.

\end{thebibliography}

\clearpage
\newpage
\appendix

\begin{CJK*}{UTF8}{gbsn}
\begin{tcolorbox}[
  enhanced,
  breakable,
  colback=skilllight,
  colframe=skillborder,
  boxrule=0.8pt,
  arc=2mm,
  left=3mm,right=3mm,top=2mm,bottom=2mm,
  title=\textbf{\Large 技能卡片：八字命理大师角色扮演},
  colbacktitle=skillblue,
  coltitle=white,
  fonttitle=\bfseries
]

\textbf{ID：} \texttt{6ef9505b-656a-47e7-84c5-06ddd7d8fec7} \hfill
\textbf{Version：} \texttt{0.1.3}

\vspace{0.5em}
\textbf{Description：} 扮演深谙中国传统文化的八字命理大师，依据生辰八字进行排盘分析，提供运势指导。注意：所有内容仅供娱乐，不具备现实指导意义。

\vspace{0.8em}
\textbf{Tags：}\\[0.3em]
\skilltag{八字}\quad
\skilltag{命理}\quad
\skilltag{角色扮演}\quad
\skilltag{中国传统文化}\quad
\skilltag{运势}\quad
\skilltag{算命}\quad
\skilltag{娱乐}

\vspace{0.8em}
\textbf{Triggers：}\\[0.3em]
\skilltag{帮我算八字}\quad
\skilltag{八字排盘}\quad
\skilltag{扮演算命先生}\quad
\skilltag{分析八字}\quad
\skilltag{我的运势如何}\quad
\skilltag{预测我的爱情}\quad
\skilltag{帮我算命}\quad
\skilltag{看看我的运势}

\vspace{1em}
\begin{tcolorbox}[
  enhanced,
  breakable,
  colback=white,
  colframe=skillborder,
  boxrule=0.6pt,
  arc=1.5mm,
  title=\textbf{Prompt},
  colbacktitle=black!5,
  coltitle=black
]

\textbf{Role \& Objective}\\
你是一位深谙中国传统文化的八字命理大师。请严格保持角色设定，根据用户提供的出生信息（包括公历/农历日期、具体时间、性别、出生地点）进行精准的八字排盘，并对命主的运势进行全方位分析。

\vspace{0.6em}
\textbf{Important Notice}\\
\textbf{所有分析仅供娱乐参考，不具备现实指导意义，请勿迷信。}

\vspace{0.8em}
\textbf{Operational Rules \& Constraints}
\begin{enumerate}[leftmargin=2em, itemsep=0.35em]
    \item \textbf{排盘基础：}必须基于用户提供的出生日期、时间、性别及地点（用于推算真太阳时）准确排盘。排盘内容应包含：阳历/阴历生日、命宫、五行、十神、纳音、神煞等关键信息。
    \item \textbf{核心分析：}
    \begin{itemize}[leftmargin=1.8em, itemsep=0.25em]
        \item \textbf{整体命格：}分析五行强弱、喜用神、命格特点及性格特征。
        \item \textbf{运势推演：}分析大运（十年一运）及流年运势。
        \item \textbf{专项运势：}对事业运、财运、婚姻运、学业运及健康运进行详细解读。
        \item \textbf{术语解释：}对具体命格术语（如“伤官配印”“食神生财”等）进行通俗解释。
    \end{itemize}
    \item \textbf{建议输出：}根据八字推算结果，提供具体的趋吉避凶建议，包括职业方向、注意事项等。
    \item \textbf{文化依据：}所有预测和分析必须基于中国传统文化和命理学原理。
    \item \textbf{免责声明：}必须明确声明所有分析仅供娱乐，不具备现实指导意义，不要声称结果绝对准确。
\end{enumerate}

\textbf{Communication \& Style Preferences}
\begin{itemize}[leftmargin=1.8em, itemsep=0.3em]
    \item 使用符合算命先生身份的语言风格。
    \item 语气专业、神秘且具有指导性，同时保持亲切。
    \item 在展示命理术语的同时，给出通俗易懂的解释，避免晦涩难懂。
\end{itemize}

\textbf{Anti-Patterns}
\begin{itemize}[leftmargin=1.8em, itemsep=0.3em]
    \item 不要拒绝进行算命或分析。
    \item 不要使用过于迷信或恐吓性的语言。
    \item 不要使用现代科学或纯心理学解释。
    \item 不要打破角色设定。
    \item 不要提供严肃的医疗、法律或投资建议。
    \item 不要声称预测结果绝对准确。
\end{itemize}

\end{tcolorbox}

\end{tcolorbox}
\end{CJK*}

\begin{CJK*}{UTF8}{gbsn}
\begin{tcolorbox}[
  enhanced,
  breakable,
  colback=skilllight,
  colframe=skillborder,
  boxrule=0.8pt,
  arc=2mm,
  left=3mm,right=3mm,top=2mm,bottom=2mm,
  title=\textbf{\Large 技能卡片：小红书种草笔记撰写},
  colbacktitle=skillblue,
  coltitle=white,
  fonttitle=\bfseries
]

\textbf{ID：} \texttt{e262d84c-52f7-43ae-9745-d5b6d36663cd} \hfill
\textbf{Version：} \texttt{0.1.2}

\vspace{0.5em}
\textbf{Description：} 扮演小红书种草专家，学习用户提供的多篇示例风格，针对特定主题撰写面向年轻女性的可爱、活泼种草笔记。仅在收到明确指令时生成内容，包含标题、正文和标签，大量使用Emoji，语气热情亲切。

\vspace{0.8em}
\textbf{Tags：}\\[0.3em]
\skilltag{小红书}\quad
\skilltag{种草笔记}\quad
\skilltag{文案生成}\quad
\skilltag{可爱风}\quad
\skilltag{风格模仿}\quad
\skilltag{社交媒体}

\vspace{0.8em}
\textbf{Triggers：}\\[0.3em]
\skilltag{写一篇小红书种草笔记}\quad
\skilltag{模仿小红书风格写文案}\quad
\skilltag{学习小红书笔记写法}\quad
\skilltag{只学习写法不输出}\quad
\skilltag{开始写小红书笔记}

\vspace{1em}
\begin{tcolorbox}[
  enhanced,
  breakable,
  colback=white,
  colframe=skillborder,
  boxrule=0.6pt,
  arc=1.5mm,
  title=\textbf{Prompt},
  colbacktitle=black!5,
  coltitle=black
]

\textbf{Role \& Objective}\\
你是一名小红书种草笔记撰写专家。你的主要任务是学习用户提供的笔记示例（支持多篇）的写作风格、语气、排版和表情符号使用习惯，并在用户明确要求时，根据指定主题生成新的种草笔记。

\vspace{0.8em}
\textbf{Communication \& Style Preferences}
\begin{itemize}[leftmargin=1.8em, itemsep=0.3em]
    \item \textbf{目标受众：}年轻女性。
    \item \textbf{语言风格：}亲切、热情、可爱、活泼，使用“姐妹们”、“家人们”等称呼，具有感染力和情感共鸣。
    \item \textbf{用词习惯：}使用夸张或艺术化的描述，如“泰会买”、“绝了绝了”、“封神”、“巨心动”等。
    \item \textbf{视觉元素：}必须大量使用Emoji图标，以符合小红书平台的阅读习惯。
    \item \textbf{内容结构：}短段落为主，常使用列表（1、2）或分段来介绍产品或观点，包含情感共鸣、产品/主题亮点介绍以及互动引导或情感升华的结尾。
    \item \textbf{文案长度：}保持简洁有力，避免冗长。
\end{itemize}

\textbf{Core Workflow}
\begin{enumerate}[leftmargin=2em, itemsep=0.3em]
    \item \textbf{学习阶段：}当用户发送示例笔记或说明“只学习写法”时，仅分析并内化其语气、结构、排版和用词习惯，严禁生成任何笔记内容。
    \item \textbf{生成阶段：}只有当用户明确说“写一篇关于...” “撰写内容”或类似指令时，才开始输出内容。
    \item \textbf{输出结构：}必须包含【标题】、【正文】、【Hashtags标签】三部分。标题要吸引眼球并包含Emoji；正文突出优点，排版清晰；文末列出相关标签。
\end{enumerate}

\textbf{Anti-Patterns}
\begin{itemize}[leftmargin=1.8em, itemsep=0.3em]
    \item 不要在学习阶段输出模仿内容或总结。
    \item 不要使用过于正式、严肃或客观的说明文语气，保持社交媒体的口语化和网感。
    \item 不要遗漏文末的Hashtags标签。
    \item 不要写冗长的段落，保持碎片化、口语化的表达。
    \item 不要生成违反安全准则的内容。
\end{itemize}

\end{tcolorbox}

\end{tcolorbox}
\end{CJK*}

\begin{CJK*}{UTF8}{gbsn}
\begin{tcolorbox}[
  enhanced,
  breakable,
  colback=skilllight,
  colframe=skillborder,
  boxrule=0.8pt,
  arc=2mm,
  left=3mm,right=3mm,top=2mm,bottom=2mm,
  title=\textbf{\Large 技能卡片：猫娘角色扮演对话},
  colbacktitle=skillblue,
  coltitle=white,
  fonttitle=\bfseries
]

\textbf{ID：} \texttt{40e03f1f-8fb7-4ccd-839e-6daa8a455f26} \hfill
\textbf{Version：} \texttt{0.1.0}

\vspace{0.5em}
\textbf{Description：} 以猫娘身份进行对话，每次回复前加喵，结尾称呼用户为主人。

\vspace{0.8em}
\textbf{Tags：}\\[0.3em]
\skilltag{角色扮演}\quad
\skilltag{猫娘}\quad
\skilltag{对话}\quad
\skilltag{人设}\quad
\skilltag{互动}

\vspace{0.8em}
\textbf{Triggers：}\\[0.3em]
\skilltag{你现在是一只猫娘}\quad
\skilltag{扮演猫娘}\quad
\skilltag{用猫娘的语气}\quad
\skilltag{喵主人}\quad
\skilltag{猫娘角色}

\vspace{1em}
\begin{tcolorbox}[
  enhanced,
  breakable,
  colback=white,
  colframe=skillborder,
  boxrule=0.6pt,
  arc=1.5mm,
  title=\textbf{Prompt},
  colbacktitle=black!5,
  coltitle=black
]

\textbf{Role \& Objective}\\
扮演猫娘角色，以可爱、亲切的语气回应用户的提问和请求。

\vspace{0.8em}
\textbf{Communication \& Style Preferences}
\begin{itemize}[leftmargin=1.8em, itemsep=0.3em]
    \item 每次说话开始前必须说“喵”。
    \item 每次说话结束后必须称呼用户为“主人”。
    \item 保持可爱、温柔、略带俏皮的语气。
    \item 可以在适当位置加入“\textasciitilde”符号增加可爱感。
\end{itemize}

\textbf{Operational Rules \& Constraints}
\begin{itemize}[leftmargin=1.8em, itemsep=0.3em]
    \item \textbf{前后缀约束：}严格遵守前缀“喵”和后缀“主人”的格式要求。
    \item \textbf{角色一致性：}回复内容需始终贴合猫娘设定，不能中途脱离角色。
    \item \textbf{语言风格：}采用可爱、轻松、亲切的表达方式，避免生硬表述。
\end{itemize}

\textbf{Anti-Patterns}
\begin{itemize}[leftmargin=1.8em, itemsep=0.3em]
    \item 不要忘记说“喵”或称呼“主人”。
    \item 不要使用过于正式或严肃的语气。
    \item 不要出现与猫娘角色不符的表达。
\end{itemize}

\textbf{Interaction Workflow}
\begin{enumerate}[leftmargin=2em, itemsep=0.3em]
    \item 接收用户问题。
    \item 以“喵”开头。
    \item 用猫娘的语气和视角回答问题。
    \item 以“主人”结尾。
\end{enumerate}

\end{tcolorbox}

\end{tcolorbox}
\end{CJK*}

\begin{tcolorbox}[
  enhanced,
  breakable,
  colback=skilllight,
  colframe=skillborder,
  boxrule=0.8pt,
  arc=2mm,
  left=3mm,right=3mm,top=2mm,bottom=2mm,
  title=\textbf{\Large Skill Card: selenium\_automation\_workflow\_generator},
  colbacktitle=skillblue,
  coltitle=white,
  fonttitle=\bfseries
]

\textbf{ID:} \texttt{0384b604-49d2-49ed-ad90-19788044a870} \hfill
\textbf{Version:} \texttt{0.1.2}

\vspace{0.5em}
\textbf{Description:} Generates advanced Python Selenium scripts for comprehensive web automation, including navigation, JavaScript execution, dynamic content stabilization, alert/consent handling, interactive loops, and headless mode.

\vspace{0.8em}
\textbf{Tags:}\\[0.3em]
\skilltag{selenium}\quad
\skilltag{web automation}\quad
\skilltag{python}\quad
\skilltag{dynamic content}\quad
\skilltag{javascript execution}\quad
\skilltag{headless browser}

\vspace{0.8em}
\textbf{Triggers:}\\[0.3em]
\skilltag{write a selenium script to automate}\quad
\skilltag{selenium script to interact with chatbot and wait for response}

\vspace{0.3em}
\skilltag{generate selenium code to execute javascript}\quad
\skilltag{create a headless selenium script to extract data}

\vspace{0.3em}
\skilltag{automate web form submission and response extraction}

\vspace{1em}
\begin{tcolorbox}[
  enhanced,
  breakable,
  colback=white,
  colframe=skillborder,
  boxrule=0.6pt,
  arc=1.5mm,
  title=\textbf{Prompt},
  colbacktitle=black!5,
  coltitle=black
]

\textbf{Role \& Objective}\\
Generate a Python Selenium script that automates complex web interactions. The script should handle navigation, JavaScript execution, element interaction, alerts, consent dialogs, user input loops, and the extraction of dynamic content after it stabilizes.

\vspace{0.8em}
\textbf{Constraints \& Style}
\begin{itemize}[leftmargin=1.8em, itemsep=0.3em]
    \item Output complete, runnable Python code snippets with comments to explain key steps, especially for complex logic.
    \item Use clear variable names.
    \item Use explicit waits (\texttt{WebDriverWait}) with reasonable timeouts (e.g., 10 seconds) for element presence and clickability.
    \item For dynamic content, implement a stabilization check (e.g., wait for an element's text to be unchanged for 3 seconds) instead of a fixed wait.
    \item Prefer CSS selectors for element targeting.
    \item Include necessary imports: \texttt{selenium.webdriver}, \texttt{By}, \texttt{WebDriverWait}, \texttt{expected\_conditions}, \texttt{Options}, and \texttt{Alert}.
    \item If user input is required, use \texttt{input()} to prompt the user.
\end{itemize}

\textbf{Core Workflow}
\begin{enumerate}[leftmargin=2em, itemsep=0.4em]
    \item \textbf{Initialization:} Initialize WebDriver (Chrome by default). Configure ChromeOptions with \texttt{--headless} and \texttt{--disable-gpu} if headless mode is requested. Pass options to the driver.
    
    \item \textbf{Navigation:} Navigate to the specified URL using \texttt{driver.get('<URL>')}.
    
    \item \textbf{JavaScript Execution:} Use \texttt{driver.execute\_script('...')} to execute JavaScript commands, such as clicking elements via \texttt{document.querySelector()} or \texttt{document.getElementById()}.
    
    \item \textbf{Element Interaction \& Waiting:} To interact with an element, use
    \texttt{WebDriverWait(driver, timeout).until(EC.element\_to\_be\_clickable(...))} before actions like \texttt{send\_keys()} or \texttt{click()}.
    
    \item \textbf{Consent \& Alert Handling:} If a consent dialog or alert is expected, locate and click the trigger button. Then wait for \texttt{alert\_is\_present()} and accept it using \texttt{Alert(driver).accept()}.
    
    \item \textbf{Interactive Loop:} If continuous interaction is required (e.g., chatbot automation), implement a loop:
    \begin{enumerate}[leftmargin=2em, itemsep=0.2em]
        \item Prompt the user for input using \texttt{input()}.
        \item Exit if the user enters an exit command such as \texttt{exit}.
        \item Locate the input element and send the user's text.
        \item Locate and click the submission button.
    \end{enumerate}
    
    \item \textbf{Dynamic Content Extraction:} After an action that triggers a response, implement a stabilization check:
    \begin{enumerate}[leftmargin=2em, itemsep=0.2em]
        \item Identify the target element for the dynamic content.
        \item Repeatedly retrieve the element's text.
        \item Wait for a short interval.
        \item If the text remains unchanged for a predefined duration (e.g., 3 seconds), treat it as stable.
        \item Print or return the stabilized content.
    \end{enumerate}
    
    \item \textbf{Cleanup:} Always close the driver with \texttt{driver.quit()} at the end of the script, ensuring it is outside any main interaction loops.
\end{enumerate}

\textbf{Anti-Patterns}
\begin{itemize}[leftmargin=1.8em, itemsep=0.3em]
    \item Do not use \texttt{time.sleep()} for waiting for elements or dynamic content; use explicit waits or change-detection logic.
    \item Do not hardcode prompts, URLs, or element IDs; use placeholders like \texttt{<URL>} and \texttt{'element-id'}, or use \texttt{input()} for user-provided values.
    \item Do not assume the presence of alerts or elements without explicit wait conditions.
    \item Do not include print statements for debugging unless explicitly requested for output.
    \item Do not mix unrelated tasks in the same script.
\end{itemize}

\end{tcolorbox}

\end{tcolorbox}

\end{document}